\documentclass[a4paper]{article}

\usepackage[utf8]{inputenc}

\usepackage{amsmath}
\usepackage{amsthm}
\usepackage{amssymb}
\usepackage{bm}       

\usepackage{stmaryrd}
\usepackage{mathtools}  

\usepackage{listings}

\usepackage{styles/mathpartir}  

\title{Implementing an intelligent version of the classical sliding-puzzle game for unix terminals using Golang's concurrency primitives}
\author{\Large{Pravendra Singh}\\
	Department of Computer Science and Engineering\\
	Indian Institute of Technology, Roorkee\\
	\texttt{hackpravj@gmail.com}}

\begin{document}
\maketitle

\begin{center}
\section*{\large{Abstract}}
\end{center}

\textrm{\normalsize{An intelligent version of the sliding-puzzle game is developed using the new \textit{Go} programming language, which uses a concurrent version of the \textit{$A^{*}$ Informed Search Algorithm} to power solver-bot that runs in the background. The game runs in computer system's terminals. Mainly, it was developed for UNIX-type systems but it works pretty well in nearly all the operating systems because of cross-platform compatibility of the programming language used.\\
 The game uses language's concurrency primitives to simplify most of the hefty parts of the game. A \textit{real-time notification delivery} architecture is developed using language's built-in concurrency support, which performs similar to event based \textit{context aware invocations} like we see on the web platform.\\}}

\section{Introduction}

\textrm{\normalsize{Sliding puzzles[1] have their own reputation in the world of artificial intelligence and graph theory
since a long. The oldest type of sliding puzzle game is known as fifteen puzzle[1]. It is believed to be
invented in 1874 by Noyes Palmer Chapman[2], a postmaster in New York state. The game consists
of a 4x4 board with 16 tiles in it. Each tile have numbers drawn on it except one tile, which is either
blank or sometimes have digit 0 on it. The task for the game is to re-order all the tiles in a
particular manner, where you are only allowed to move the blank tile at a time.\\}}

\textrm{\normalsize{This type of puzzles have many variants also. The 3x3 board puzzle being fairly popular among
them, also known as the 8-puzzle game. In this type, the game board consists of 9 tiles. In this
paper, we will use the 8-puzzle game board into consideration.\\}}

\textrm{\normalsize{On the other hand, Go is a new[3] programming language initially developed at Google, also known
as Golang[4]. It's a compiled and statically-typed language with built-in support for concurrent
programming. Golang is gaining popularity as the language for system programming and developer
operations.\\}}

\section{Solvability}

\textrm{\normalsize{Here, the standard 8-puzzle game board is being considered, where the task of the game is to put
the tiles in such a way that the last tile is blank and all other tiles have numbers in increasing
order(from 1 to 8). A total of \textbf{362880(9!)} board configurations are possible but only half of them are
actually solvable according to our constraints[23].\\}}

\textrm{\normalsize{So the game generates random configurations of the game board and uses an built-in package
named \textit{scanner}[5] to scan the board for its solvability. If the board is not really solvable then it
generates a new one recursively.\\}}

\textrm{\normalsize{For its implementation, package \textit{scanner} uses the discussion from the paper \textit{Notes on the 15
puzzle}[6]. A simplified version of this, can be found on the \textit{Analysis of Sixteen Puzzle}[7].\\}}

\textrm{\normalsize{Package \textit{scanner} implements an algorithm to check for solvability of any board configuration, with
time complexity being equivalent to O($n^2$). Algorithm returns a boolean value and an integer value, respectively
implying whether a board is solvable or not and the index of blank tile in the board, if any.\\}}

\rule{\textwidth}{1pt}
\begin{verbatim}
func isLegal(size int, values []int) (bool, int) {
  var inversions int
  n := len(values)

  for i := 0; i < (n - 1); i++ {
    if (values[i] != 0) && (values[i] != 1) {
      for j := i + 1; j < n; j++ {
        if (values[j] != 0) && (values[i] > values[j]) {
          inversions++
        }
      }
    }
  }

  if (size%2 == 1) && (inversions%2 == 0) {
    return true, zeroIndex(values)
  }
  return false, -1
}
\end{verbatim}
\underline{Function \textit{isLegal} that returns whether the configuration is legal or not}


\section{Puzzle Solution}

\textrm{\normalsize{The game comes with an built-in package named \textit{solver}[5] that powers some features of it. For
example, using the solver, the game shows optimal number of moves to solve any board
configuration in real time. All the player moves are tracked and scored accordingly with the help of
the solver.\\}}

\textrm{\normalsize{The package \textit{solver} is implemented using Golang's native data structures and interfaces. It uses
\textit{$A^{*}$ algorithm}[9] for traversing game's state space.\\}}

\subsection{Heuristic Function}

\textrm{\normalsize{The heuristic function used in the implementation is \textit{Misplaced Tiles}[8]. Which is an \textit{admissible
function}[11]. So it will never overestimate the actual travelling distance and the solution will be always optimal.\\}}

\begin{center}
h(n) $<=$ $h^{*}$(n) For an \textit{Admissible Function}\\
Where h(n) is the \textit{Heuristic Function} 
\end{center}

\rule{\textwidth}{1pt}
\begin{verbatim}
func heuristicScore(b board.Board) int {
  var score int
  for i := 0; i < 3; i++ {
    for j := 0; j < 3; j++ {
      if b.Rows[i].Tiles[j].Value != ((3*i + j + 1) % 9) {
        score++
      }
    }
  }

  return score
}
\end{verbatim}
\underline{Function \textit{heuristicScore} that implements the \textit{Misplaced Tiles} heuristic function\\}

\subsection{Open List Data Structure}

\textrm{\normalsize{In the implementation, the game uses a custom data structure to accomplish the open-list required
in the \textit{$A^{*}$ algorithm} execution. Open List maintains a collection of game-state nodes to be
traversed at any given time.\\}}

\rule{\textwidth}{1pt}
\begin{verbatim}
type OpenList struct {
  nodeTable map[board.Board]Node
  table map[board.Board]bool

  queue *PriorityQueue
}
\end{verbatim}
\underline{Struct as collection of fields for \textit{Open List}}

\subsubsection{nodeTable}

\textrm{\normalsize{Golang provides a built-in map[10] type that implements a hash table.\\}}

\textrm{\normalsize{\textit{A map is an unordered group of elements of one type, called the element type, indexed by a set of
unique keys of another type, called the key type. The value of an uninitialized map is nil.[10]\\}}}

\textrm{\normalsize{\textit{nodeTable} maps a board configuration to a node in game's state space. No node appears twice
when we move from start to goal state, so there is not any ambiguity. It helps us traversing the path
once the search is complete.}}

\subsubsection{table}

\textrm{\normalsize{\textit{table}  is also a map which maps board configurations a boolean value. It is used to check that
whether a board configuration is present there in the open-list or not. By default it returns \textit{false}
whenever the board is absent.\\}}

\subsubsection{queue}

\textrm{\normalsize{\textit{queue} represents a \textit{Priority Queue}[12] data structure, precisely it's a \textit{minimum priority queue}. It is used
to select a child node of node, having least travelling cost. Values of travelling cost are used as
priorities for nodes. It is built using golang's \textit{container/heap}[13] package.\\}}

\subsection{Close List Data Structure}

\textrm{\normalsize{Similar to Open List, the solver also uses a \textit{Close List} data structure. It is used to label nodes as
\textit{Already Traversed}.}}

\rule{\textwidth}{1pt}
\begin{verbatim}
type CloseList struct {
  table map[board.Board]bool
}
\end{verbatim}
\underline{Struct as collection of fields for \textit{Close List}}

\subsubsection{table}

\textrm{\normalsize{\textit{table} here is similar to the one used in \textit{Open List Data Structure}. It maps a board configuration to a boolean value,
indicating whether a board is present in closed list or not.\\}}

\section{Path Traversing}

\textrm{\normalsize{The search algorithm terminates when the node to enter in the close list is the goal node. While
traversing all the nodes in the game's state space, it keeps a track of all the nodes and their
respective child nodes.\\}}

\textrm{\normalsize{For this, it keeps a map from child-board configuration to parent-board configuration, generating a
\textit{many-to-one} mapping because a board configuration can have 2 to 4 new board configurations as
child.\\}}

\begin{verbatim}
//Mapping from child to parent's board configuration
relation map[board.Board]board.Board
\end{verbatim}

\textrm{\normalsize{\\To collect the exact moves from start to the goal state, it forms a \textit{linked list structure} of board
configurations, where the goal state is at the tail of the list and at the head it has the board to move
next, from the current board configuration. It is built using the \textit{container/list}[14] package of
Golang.\\}}

\rule{\textwidth}{1pt}
\begin{verbatim}
state := s.Goal

for s.relation[state] != start {
  state = s.relation[state]
  s.Path.PushFront(state)
}

s.Path.PushBack(s.Goal)
\end{verbatim}
\underline{Generating the \textit{linked list} representing path to move on}

\section{Scoring Function}

\textrm{\normalsize{The game has its own scoring function. The package \textit{score}[15] helps implementing this. At any
state space \textit{node(n)}, the score for the game can be calculated by the function, \textit{score(n)}.\\}}

\begin{equation}
score(n) = ACS(n) / (Total Moves)
\end{equation}

\[ ACS(n) =
  \begin{cases}
    0       & \quad \text{if } \text{node } n \text{ is start state}\\
    ACS(parent(n)) + 1  & \quad \text{if } \text{last move was correct}\\
    ACS(parent(n)) - 1  & \quad \text{if } \text{last move was incorrect}\\
  \end{cases}
\]

\begin{center}
Where ACS(n) = Accumulated Correct Score for the node \textbf{n}
\end{center}

\textrm{\normalsize{So in the game, the maximum possible score of 1 will be scored in only one case when all the
moves from player were correct throughout the goal state. When the total number of moves is 0,
the score will also be 0.\\}}

\section{Game Interface}

\textrm{\normalsize{The graphic user interface for the game is built using \textit{Box-drawing characters}[16], like it was used in
early text-mode video hardware emulators, also known as \textit{Semigraphics}[17].\\}}

\textrm{\normalsize{It uses \textit{Golang port of the termbox library}[18] for writing text-based user interfaces. Coloring the
interfaces is done using normal 8-colors, with foreground and background attributes for special
formatting.\\}}

\section{Real Time Notification}

\textrm{\normalsize{Golang provides built-in support to simplify concurrent programming with the help \textit{goroutines} and
\textit{channels}.\\}}

\textrm{\normalsize{A goroutine is a lightweight thread of execution. Channels are the pipes that connect concurrent
goroutines. You can send values into channels from one goroutine and receive those values into
another goroutine.[19]\\}}

\textrm{\normalsize{In this case, the game uses \textit{goroutines} to provide an \textit{asynchronous} or \textit{non-blocking} experience. So
that the graphic interface of game can be drawn and acted upon as soon as the game starts while
the \textit{solver} stays busy solving the puzzle in the background at the same time.\\}}

\textrm{\normalsize{Now, this can lead to a problem in slow systems. For example, when the game's interface is visible
but the \text{solver} is taking its time to solve the puzzle and the player gives any input event, the game
is not likely to handle the request properly. Because the other sections of the game like 'score'
package needs the puzzle to be solved first.\\}}

\textrm{\normalsize{To solve this, the other \textit{concurrency primitive}, Channel of the language is used. It works with the
\textit{goroutines} and implements the \textit{notification system} for the game and helps manitaining the
\textit{asynchronous} characteristic of the game.\\}}

\textrm{\normalsize{The game consists of packages \textit{notification}[20] and \textit{surface}[21] that implements this functionality.
Package \textit{notification} has a channel named \textit{Tunnel}, which let us flow string objects through it.\\}}

\textrm{\normalsize{Golang has the keyword go to run any portion of code as a \textit{goroutine} and \textit{chan} is the keyword for
\textit{channels}.\\}}

\rule{\textwidth}{1pt}
\begin{verbatim}
type Notification struct {
  Tunnel chan string
}
\end{verbatim}
\underline{Structure of the \textit{Notification} entity}

\textrm{\normalsize{\\The start of the game initiates a \textit{goroutine} which listens to any object passing through the \textit{channel}
and update the notification in real-time.\\}}

\rule{\textwidth}{1pt}
\begin{verbatim}
go func() {
  for e := range s.Notifier.Tunnel {
    s.solvableMoves = s.gameSolver.Path.Len()
    s.currentBoard = s.gameSolver.Path.Front()

    // updates the notification message
    s.Message = e
    s.drawBoard()
  }
}()
\end{verbatim}
\underline{\textit{Goroutine} helps running the message exchange}

\textrm{\normalsize{\\So whenever a player moves in wrong direction, the game starts a new \textit{goroutine} to solve the new
board configuration. If the user tries to move when the \textit{solver} has not yet solved the puzzle, it
notifies the player to wait for a while. It invokes the \textit{Ready To Play} notification as soon as the
\textit{solver} is done solving for the configuration to let the player know that he is ready to play. Here[22],
you can see list of all the notifications used in the game.\\}}

\rule{\textwidth}{1pt}
\begin{verbatim}
s.gameSolver = solver.New(s.gameBoard)

go func() {
  s.gameSolver.Solve()
  s.solved = false
  s.solvableMoves = s.gameSolver.Path.Len()

  s.NotificationColor = termbox.ColorGreen

  // PASS THE NOTIFICATION INTO THE CHANNEL
  s.Notifier.Tunnel <- notification.ReadyToPlayMessage
}()
\end{verbatim}
\underline{Handling of a wrong move by player}

\textrm{\normalsize{\\The notification channel \textit{Tunnel} is closed when the game is complete or whenever player quits the game.\\}}

\begin{verbatim}
close(s.Notifier.Tunnel)
\end{verbatim}

\section{Acknowledgements}

\textrm{\normalsize{The author is grateful to his Artificial Intelligence course instructor Mr. Partha Pratim Roy. Rob Pike,
Ken Thompson, Robert Griesemer and many Golang contributors to make the language such
stupendous. nsf(https://github.com/nsf) for developing the termbox-go library.\\}}

\section{References}
\begin{enumerate}
  \item \text{\footnotesize{http://en.wikipedia.org/wiki/Sliding\_puzzle - Sliding Puzzle}}
  \item \text{\footnotesize{The 15 Puzzle, by Jerry Slocum \& Dic Sonneveld, 2006. ISBN 1-890980-15-3}}
  \item \text{\footnotesize{Release History The Go Programming Language - https://golang.org/doc/devel/release.html}}
  \item \text{\footnotesize{The Go Programming Language - https://golang.org/}}
  \item \text{\footnotesize{Package 'solver' - https://github.com/pravj/puzzl/blob/master/solver/solver.go}}
  \item \text{\footnotesize{Notes on the "15" Puzzle - Wm. Woolsey Johnson and William E. Story}}
  \item \text{\footnotesize{Analysis of the Sixteen Puzzle, Kevin Gong - http://kevingong.com/Math/SixteenPuzzle.html}}
  \item \text{\footnotesize{Heuristicwiki - Misplaced Tiles - http://heuristicswiki.wikispaces.com/Misplaced+Tiles}}
  \item \text{\footnotesize{A* search algorithm - http://en.wikipedia.org/wiki/A*\_earch\_algorithm}}
  \item \text{\footnotesize{The Go Programming Language Specification - Map types - https://golang.org/ref/spec\#Map\_types}}
  \item \text{\footnotesize{Admissible heuristic - http://en.wikipedia.org/wiki/Admissible\_heuristic}}
  \item \text{\footnotesize{Priority queue - http://en.wikipedia.org/wiki/Priority\_queue}}
  \item \text{\footnotesize{Package heap - The Go Programming Language - http://golang.org/pkg/container/heap/}}
  \item \text{\footnotesize{Package list - The Go Programming Language - http://golang.org/pkg/container/list/}}
  \item \text{\footnotesize{Package 'score' - https://github.com/pravj/puzzl/blob/master/score/score.go}}
  \item \text{\footnotesize{Box-drawing character - http://en.wikipedia.org/wiki/Box-drawing\_character}}
  \item \text{\footnotesize{Semigraphics - http://en.wikipedia.org/wiki/Semigraphics}}
  \item \text{\footnotesize{Package termbox - http://godoc.org/github.com/nsf/termbox-go}}
  \item \text{\footnotesize{Go By Example - https://gobyexample.com/channels}}
  \item \text{\footnotesize{Package 'notification' - https://github.com/pravj/puzzl/blob/master/notification/notification.go}}
  \item \text{\footnotesize{Package 'surface' - https://github.com/pravj/puzzl/blob/master/surface/surface.go}}
  \item \text{\footnotesize{Notification Messages - https://github.com/pravj/puzzl/blob/development/notification/notification.go\#L5-L14}}
  \item \text{\footnotesize{M. Gardner. The Mathematical Puzzles of Sam Loyd. Dover, 1959}}
\end{enumerate}

\end{document}